\documentclass[runningheads]{llncs}

\usepackage{graphicx,url}
\usepackage[utf8]{inputenc}

\usepackage{cmap}
\usepackage{mathtools}
\usepackage{times}
\usepackage{lastpage}
\usepackage{indentfirst}
\usepackage{color}
\usepackage{caption}
\usepackage{subcaption}
\usepackage{makecell}
\usepackage{longtable}
\usepackage{stmaryrd}
\usepackage{amssymb}
\usepackage{acronym}
\usepackage{fancyhdr}
\usepackage{algorithm}
\usepackage{colortbl}
\usepackage{upgreek}
\usepackage{amsmath}
\usepackage{mathrsfs}
\usepackage{placeins}
\usepackage{blindtext}
\usepackage{todonotes}
\usepackage{pdfpages}
\usepackage{dirtytalk}
\usepackage{bm}
\usepackage{bbm}
\usepackage{relsize}
\usepackage{todonotes}

\usepackage{mathtools}
\DeclarePairedDelimiter\norm{\big\lVert}{\big\rVert}%

\usepackage{algorithm,algpseudocode}
\usepackage{comment}

\newcommand*{\pd}[3][]{\ensuremath{\frac{\partial^{#1} #2}{\partial #3}}}

\newcommand{\by}{\hspace{-0.5 pt}\mathsmaller{\times}\hspace{-0.5 pt}}

\newcommand{\mb}{ \mathbf}

\newcommand{\LGCLVOAutoL}{$\texttt{LGC\textunderscore LVO\textunderscore AutoL}$ }
\newcommand{\LGCLVOAuto}{$\texttt{LGC\textunderscore LVO\textunderscore Auto}$ }
\newcommand{\LGCLVOAutoD}{$\texttt{LGC\textunderscore LVO\textunderscore AutoD}$ }

\newcommand{\Ln }{\mb{L}_{\mathbb{N}}}
\newcommand{\Lc }{\mb{L}_{\mathbb{C}}}
\newcommand{\La }{\mb{L}}

\newcommand{\Ma}{\mathbb{R}}
\newcommand{\lab}{\mathcal{L}}
\newcommand{\unlab}{\mathcal{U}}

\begin{document}
\titlerunning{Optimizing Diffusion Rate and Label Reliability in a GSSL
Classifier}
\title{Optimizing Diffusion Rate and Label Reliability in a Graph-based Semi-supervised
Classifier}

%
%
\author{Bruno Klaus de Aquino Afonso \inst{1}\orcidID{0000-0003-2086-1054} \and
Lilian Berton \inst{1}\orcidID{0000-0003-1397-6005} }
%
%
\institute{University of São Paulo, Institute of Science and Technology
\\
\email{\{bruno.klaus,lberton\}@unifesp.br}
}
\newcommand\blfootnote[1]{%
  \begingroup
  \renewcommand\thefootnote{}\footnote{#1}%
  \addtocounter{footnote}{-1}%
  \endgroup
}

\maketitle              
\begin{abstract}
Semi-supervised learning has received attention from researchers, as it allows one to exploit the structure of unlabeled data to achieve competitive classification results with much fewer labels than supervised approaches. The Local and Global Consistency (LGC) algorithm is one of the most well-known graph-based semi-supervised (GSSL) classifiers.  Notably, its solution can be written as a linear combination of the known labels. The coefficients of this linear combination depend on a parameter $\alpha$, determining the decay of the reward over time when reaching labeled vertices in a random walk. In this work, we discuss how removing the self-influence of a labeled instance may be beneficial, and how it relates to leave-one-out error. Moreover, we propose to minimize this leave-one-out loss with automatic differentiation. Within this framework, we propose methods to estimate label reliability and diffusion rate. Optimizing the diffusion rate is more efficiently accomplished with a spectral representation. Results show that the label reliability approach competes with robust $\ell_1$-norm methods and that removing diagonal entries reduces the risk of overfitting and leads to suitable criteria for parameter selection.\blfootnote{This study was financed in part by the Coordenação de Aperfeiçoamento de Nível Superior - Brasil (CAPES) - Finance Code 001, and São Paulo Research Foundation (FAPESP) grant \#18/01722-3 }

\end{abstract}

\section{Introduction}
\label{subsec:theory_ml}
\par \textit{Machine learning} (ML) is the subfield of Computer Science that aims to make a computer learn from data \cite{Mitchell97}. The task we'll be considering is the problem of \textit{classification}, which requires the prediction of a  discrete label corresponding to a \textit{class}. 

\par Before the data can be presented to an ML model, it must be represented in some way. Our input is nothing more than a collection of $n$ examples. Each object of this collection is called an \textit{instance}. For most practical applications, we may consider an instance to be a vector of $d$ dimensions. Graph-based semi-supervised learning (GSSL) relies a lot on matrix representations, so we will be representing the observed instances as an \textit{input matrix} 
\begin{equation}
    \mb{X} \in \Ma^{n\by d}.
\end{equation}
A ML classifier should learn to map an input instance to the desired output. To do this, it needs to know the \textbf{labels} (i.e. the output) associated with instances contained in the training set. In  \textbf{semi-supervised learning}, only some of the labels are known in advance \cite{Chapelle_etal_2010}. Once again, it is quite convenient to use a matrix representation, referred to as the \textit{(true) label matrix}
\begin{align}
\mb{Y}_{ij} = \begin{cases}
1 &\text{ if the $i$-th instance is associated with the $j$-th class}\\
0 &\text{otherwise}
\end{cases}
\end{align}

 \par Most approaches are \textit{inductive} in nature so that we can predict the labels of instances not seen before deployment. However, most GSSL methods are  \textit{transductive}, which simply means that we are only interested in the fixed but unknown set of labels corresponding to unlabeled instances \cite{zhou2004learning}. Accordingly, we may represent this with a \textit{classification matrix}:
\begin{align}
\mb{F} \in \Ma^{n\by c}
\end{align}

In order to separate the labeled data $\lab$ from the unlabeled data $\unlab$, we divide our matrices as following:
\begin{align}
    \mb{X} &= \left[ \mb{X_\lab}^\top, \mb{X_\unlab}^\top \right] ^\top\\
    \mb{Y} &= \left[ \mb{Y_\lab}^\top, \mb{Y_\unlab}^\top \right] ^\top\\
    \mb{F} &= \left[ \mb{F_\lab}^\top, \mb{F_\unlab}^\top \right] ^\top
\end{align}
The idea of SSL is appealing for many reasons. One of them is the possibility to
integrate the toolset developed for unsupervised learning. Namely, we may use unlabeled data to measure the density $P(\mb{x})$ within our $d$-dimensional input space. Once that is achieved, the only thing left is to take advantage of this information. To do this, we have to make use of \textit{assumptions} about the relationship between the input density $P(\mb{x})$ and the conditional class distribution $P(y\!\mid\!\mb{x})$. If we are not assuming that our datasets satisfy any kind of assumption, SSL can potentially cause a significant decrease compared to baseline performance  \cite{van2020survey}. This is currently an active area of research:  \textit{safe semi-supervised learning} is said to be attained when SSL never performs worse than the baseline, for any choice of labels for the unlabeled data. This is indeed possible in some limited circumstances, but also provably impossible for others, such as for a specific class of margin-based classifiers \cite{krijthe2018robust}.
\par
\begin{figure}[h]
\centering
\begin{subfigure}{0.35\textwidth}
  \centering
  \includegraphics[trim={0 0 0 3cm},clip,width=\linewidth]{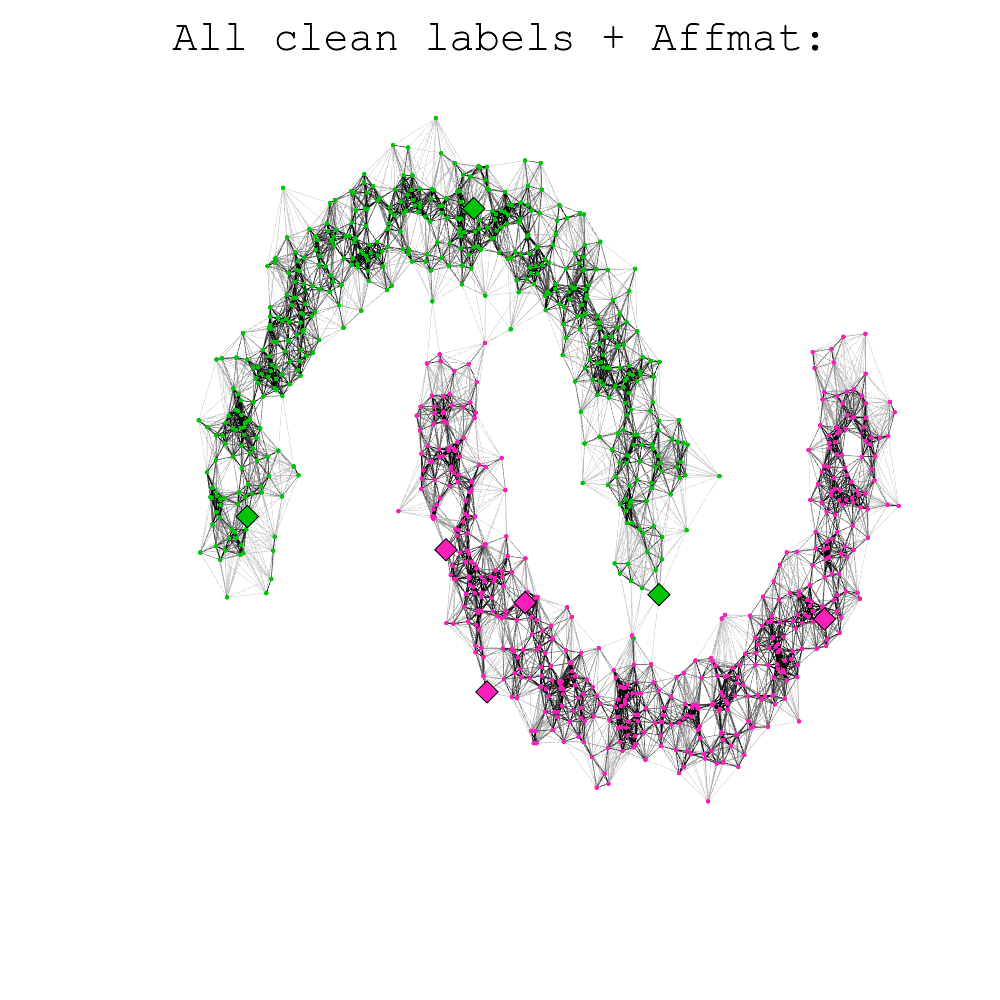}    \caption{Ground truth}
  \label{fig:ssl_ground_truth}
\end{subfigure}%
\begin{subfigure}{0.35\textwidth}
  \centering

  \includegraphics[trim={0 0 0 3cm},clip,width=\linewidth]{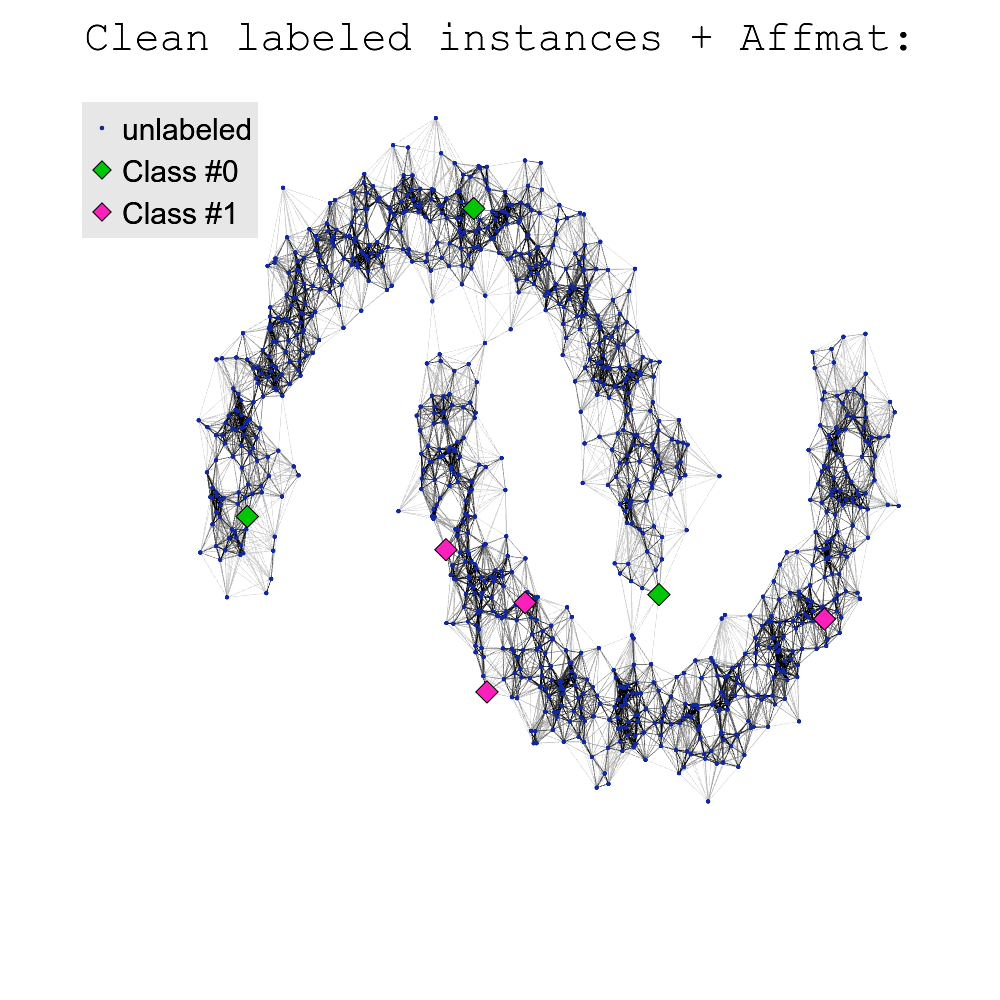}    \caption{Observed labels}
  \label{fig:ssl_observed_labels}
\end{subfigure}
\caption{An ideal scenario for semi-supervised learning}
\label{fig:ssl_ideal}
\end{figure}

 In Figure \ref{fig:ssl_ideal}, we illustrate which kind of dataset is suitable for GSSL. There are two clear spirals, one corresponding to each class. In a bad dataset for SSL, we can imagine the spiral structure to be a red herring, i.e. something misguiding. A very common assumption for SSL is the \textit{smoothness assumption}, which is one of the cornerstones for GSSL classifiers. It states that ``If two instances $\mb{x}_1$, $\mb{x}_2$ in a high-density region are close, then so should be the corresponding outputs $y_1$, $y_2$'' \cite{Chapelle_etal_2010}.
 Another important assumption is that the (high-dimensional) data lie (roughly) on a low-dimensional manifold, also known as the \textit{manifold assumption} \cite{Chapelle_etal_2010}.
If the data lie on a low-dimensional manifold, the local similarities will approximate the manifold well. As a result, we can, for example, increase the resolution of our image data without impacting performance.

Graph-based semi-supervised learning has been used extensively for many different applications across different domains. In computer vision, these include car plate character recognition \cite{catunda2019car} and  hyperspectral image classification \cite{shao2017probabilistic}. GSSL is particularly appealing if the underlying data has a natural representation as a graph. As such, it has been a promising approach for drug property prediction from the structure of molecules \cite{kearnes2016molecular}. Moreover, it has had much application in knowledge graphs, such as the development of web-scale recommendation systems \cite{ying2018graph}.

 \par GSSL methods put a greater emphasis on using \textit{geodesics} by expressing \textit{connectivity} between instances through the creation of a graph.  Many successful deep semi-supervised approaches use a similar yet slightly weaker assumption, namely that \textbf{small perturbations} in input space should cause little corresponding perturbation on the output space \cite{miyato2018virtual}.

\par It turns out that we can best express our concepts by defining a measure of \textit{similarity}, instead of distance. In particular, we search for an  \textbf{affinity matrix} $\mb{W} \in \Ma^{n \by n}$, such that 
 \begin{equation}
    \mb{W}_{ij} = \begin{cases}
    w(\mb{x}_i,\mb{x}_j)\in \mathbb{R} &\text{if $\mb{x}_i$ and $\mb{x}_j$ are considered neighbors}\\
    0 &\text{otherwise}
    \end{cases}
 \end{equation}
  where $w$ is some function determining the similarity between any two instances $\mb{x}_i,\mb{x}_j$. When constructing an affinity matrix in practice, instances are not considered neighbors of themselves, i.e. we have $\forall i \in \{1..n\}: \mb{W}_{ii} = 0$.
 
 The specification of an affinity matrix is a necessary step for any GSSL classifier, and its sparsity is often crucial for reducing computational costs. There are many ways to choose a neighborhood. Most frequently, it is constructed by looking at the \textbf{K-Nearest neighbors (KNN)} of a given instance.

One last important concept to GSSL is that of the graph Laplacian operator. This operator is analogue to the Laplace-Beltrami operator on manifolds. There are a few graph Laplacian variants, such as the \textbf{combinatorial Laplacian}
\begin{equation}
    \label{def:graph_lapl}
\Lc  = \mb{D} - \mb{W}
\end{equation}
where $\mb{D}$, called the \textit{degree matrix}, is a diagonal matrix whose entries are the sum of each row of $\mb{W}$. There is also the \textbf{normalized Laplacian}, whose diagonal is the \textit{identity matrix} $\mathbf{I}$:
\begin{equation}
    {\Ln  = \mb{I} - \mb{D}^{-\frac12}\mb{W}\mb{D}^{-\frac12} = \mb{D}^{-\frac12}\Lc \mb{D}^{-\frac12}}
    \label{eqn:sym_lap}
\end{equation}
Each graph Laplacian $\La$ induces a measure of \textbf{smoothness with respect to the graph} on a given classification matrix $\mb{F}$, namely
\begin{equation}
    {\widetilde{S}_\La}(\mb{F}) = \frac12 \sum_{k=1}^{c} (\mb{F}_{[:,k]})^{\top}\La (\mb{F}_{[:,k]}) = \frac12 tr(\mb{F}^{\top}\La \mb{F})
\end{equation}
where $tr$ is the trace of the matrix and $c$ the number of classes. If we consider each column $\mb{f}$ of $\mb{F}$ individually, then we can express graph smoothness of each graph Laplacian as
\begin{align}
    \label{def:graph_local_sm}
    \mb{f}^{\top}\Lc \mb{f} &=  \sum_{1 \leq i,j \leq n} \mb{W}_{ij} (\mb{f}_i - \mb{f}_j)^2 \\
    \mb{f}^{\top}\Ln \mb{f} &= \sum_{1 \leq i,j \leq n} \left( \frac{\mb{W}_{ij}}{\sqrt\mb{D}_{ii}} \mb{f}_i -  \frac{\mb{W}_{ij}}{\sqrt\mb{D}_{jj}}\mb{f}_j\right)^2 
\end{align}
Each graph Laplacian also has an eigendecomposition:  \begin{equation}
    \mb{L} = \mb{U}\mb{\Lambda}\mb{U}^\top
\end{equation}
where the set of columns of $\mb{U}$ is an orthonormal basis of eigenfunctions, with $\mb{\Lambda}$ a diagonal matrix with the eigenvalues.  As $\mb{U}$ is a unitary matrix, any real-valued function on the graph may be expressed as a linear combination of eigenfunctions. We map a function to the spectral domain by pre-multiplying it by $\mb{U}^\top$ (also called \textbf{graph Fourier transform}, also known as GFT). Additionally, pre-multiplying by $\mb{U}$ gives us the inverse transform. This spectral representation is very useful, as eigenfunctions that are smooth with respect to the graph have smaller eigenvalues. Outright restricting the amount of eigenfunctions is known as \textbf{smooth eigenbasis pursuit}\cite{gong2017learning}, a valid strategy for semi-supervised regularization.
\par In this work, we explore the problem of parameter selection for the Local and Global Consistency model. This model yields a propagation matrix, which itself depends on a fixed parameter $\alpha$ determining the diffusion rate. We show that, by removing the diagonal in the propagation matrix used by our baseline, a leave-one-out criterion can be easily computed. Our first proposed algorithm attempts to calculate label reliability by optimizing the label matrix, subject to constraints. This approach is shown to be competitive with robust $\ell_1$-norm classifiers. Then, we consider the problem of optimizing the diffusion rate. Doing this in the usual formulation of LGC is impractically expensive. However, we show that the spectral representation of the problem can be exploited to easily solve an approximate version of the problem. Experimental results show that minimizing leave-one-out error leads to good generalization.
\par The remaining of this paper is organized as follows. Section \ref{sec:related_work} summarizes the key ideas and concepts that are related to our work. Section \ref{sec:proposal} presents our two proposed algorithms: $\LGCLVOAutoL$ for determining label reliability, and $\LGCLVOAutoD$ for determining the optimal diffusion rate parameter. Our methodology is detailed in Section \ref{sec:methodology}, going over the basic framework and baselines. The results are presented in Section \ref{sec:results}. Lastly, concluding remarks are found in Section \ref{sec:conclusion}.

\section{Related work} \label{sec:related_work}
In this section, we present some of the algorithms and concepts that are central to our approach. We describe the inner workings of our baseline algorithm, and how eliminating diagonal entries of its propagation matrix may lead to better generalization.

\subsection{Local and Global Consistency}
\label{subsec:LGC}
The \textit{Local and Global Consistency} (LGC) \cite{zhou2004learning} algorithm is one of the most widely known graph-based semi-supervised algorithms. It minimizes the following cost:
\begin{equation}
\label{eqn:lgc_cost}
    \mathcal{Q}(\mb{F}) = \frac12 \left(tr(\mb{F}^{\top} \Ln  \mb{F}) + \mu \norm{\mb{F} - \mb{Y}}^2 \right)
\end{equation}
 LGC addresses the issue of label reliability by introducing the parameter $\mu \in (0, \infty)$. This parameter controls the trade-off between fitting labels, and achieving high graph smoothness. 
\par LGC has an analytic solution. To see this, we take the partial derivative of the cost with respect to $\mb{F}$:
\begin{align}
    \pd{\mathcal{Q}}{\mb{F}} &=  \frac12 \pd{tr(\mb{F}^{\top} \Ln  \mb{F})}{\mb{F}} + \frac12 \mu\pd{ \norm{\mb{F} - \mb{Y}}^2}{\mb{F}}\\
    &= ((1+\mu) \mb{I} - \mb{S})\mb{F} - \mu \mb{Y}
\end{align}
where $\mb{S} = \mb{D}^{-\frac12}\mb{W}\mb{D}^{-\frac12} = \mb{I} - \Ln $. By dividing the above by ($1+ \mu$), we observe that this derivative is zero exactly when
\begin{equation}
    (I - \alpha \mb{S})\mb{F} = \beta \mb{Y}
\end{equation}
with 
\begin{equation}
    \alpha = \frac{1}{1+\mu} \in (0,1)
    \label{eq:alpha}
\end{equation}
and
\begin{equation}
    \beta = 1 - \alpha 
\end{equation}
The matrix $(\mb{I} - \alpha \mb{S})$ can be shown to be positive-definite and therefore invertible, so the optimal $\mb{F}$ can be obtained as 
\begin{equation}
    \mb{F} = \beta (I - \alpha \mb{S})^{-1}\mb{Y}
\end{equation}
We hereafter refer to  $(I - \alpha S)^{-1}$ as the \textbf{propagation matrix} $\mb{P}$. Each entry $\mb{P}_{ij}$ represents the amount of label information from $X_j$ that $X_i$ inherits.  
It can be shown that the inverse is a result of a diffusion process, which is calculated via iteration:
\begin{align}
    &F(0) = \mb{Y}\\
    &F(t\!+\!1) = \alpha \mb{S} F(t) + (1\!-\!\alpha)\mb{Y} 
    \label{eqn:F_iter}
\end{align}

\noindent Moreover, it can be shown that the closed expression for $F$ at any iteration is
\begin{equation}
    F(t) = (\alpha \mb{S})^{t-1}\mb{Y} + (1-\alpha)\sum_{i=0}^{t-1}(\alpha \mb{S})^{i}\mb{Y}
\end{equation}
S is similar to $D^{-1}W$, whose eigenvalues are always in the range $[-1,1]$ \cite{zhou2004learning}. This ensures the first term vanishes as $t$ grows larger, whereas the second term converges to $\mb{P}Y$. Consequently, $\mb{P}$ can be characterized as
\begin{align}
\label{eqn:def_P}
    \mb{P} &= (1-\alpha)\lim_{t \rightarrow \infty} \sum_{i=0}^{t}(\alpha \mb{S})^{i}\\
    &= (1-\alpha)\lim_{t \rightarrow \infty} \sum_{i=0}^{t} \alpha^i \mb{D}^{\frac12}(\mb{D}^{-1}\mb{W})^{i}\mb{D}^{-\frac12}
    \label{eqn:P_iter}
\end{align}
The \textbf{transition probability matrix} $\widetilde{W} =\mb{D^{-1}W}$ makes it so we can interpret the process as a random walk. Let us imagine a particle walking through the graph according to the transition matrix. Assume it began at a labeled vertex $v_a$, and at step  $i$ it reaches a labeled vertex $v_b$, initially labeled with class $c_b$. When this happens, $v_a$ receives a \textbf{confidence boost} to class $c_b$. Alternatively, one can say that this boost goes to the entry which corresponds to the contribution from $v_b$ to $v_a$, i.e. $\mb{P}_{ab}$. This boost is proportional to $\alpha^i$. This gives us a good intuition as to the role of $\alpha$. More precisely, the contribution of vertices found later in the random walk decays exponentially according to $\alpha^i$.

\subsection{Self-influence and leave-one-out-error}
\par There is one major problem with LGC's solution: the diagonal of $\mb{P}$. At first glance, we would think that ``fitting the labels '' means looking for a model that explains our data very well. In reality,  this translates to memorizing the labeled set. The main problem resides within the \textbf{diagonal} of the propagation matrix. Any entry $\mb{P}_{ii}$ stores the \textbf{self-influence} of a vertex, which is calculated according to the expected reward obtained by looping around and visiting itself. The optimal solution w.r.t. label fitting occurs when $\alpha$ tends to zero. For labeled instances, an initial reward is given for the starting vertex itself, and the remaining are essentially ignored.

\par We argue that the diagonal is directly related to \textbf{overfitting}. It essentially tells the model to rely on the label information it knows. There are a few analogies to be made: say that we are optimizing the number of neighbors $k$ for a KNN classifier. The analog of ``LGC-style optimal label fitting'' would be to include each labeled instance as a neighbor to itself, and set $k = 1$. This is obviously not a good criterion. The answer that maximizes a \textit{proper} ``label fitting'' criteria, in this case,  is selecting $k$ that minimizes classification error with the extremely important caveat:  directly using each instance's own label is prohibited. 

\par In spite of the problems we have presented, the family of LGC solutions remains very interesting to consider. We will have to eliminate diagonals, however. Let
\begin{equation}
    \mb{H}(\alpha) := \left(\mb{(\mb{I}-\alpha \mb{S})^{-1}} - diag((\mb{I}-\alpha \mb{S})^{-1})\right)_\lab \mb{Y}_\lab
\end{equation}
By eliminating the diagonal, we obtain, for each label, its classification if it were not included in the label propagation process. As such, it can be argued that minimizing
\begin{equation}
\label{eqn:loss_H}
    \norm{\mb{H}(\alpha) - \mb{Y}_\lab}^2
\end{equation}
also minimizes the \textit{leave-one-out} (LOO)  error. There is an asterisk: each instance is still used as unlabeled data, but this effect should be insignificant. It is also interesting to \textbf{row-normalize} (a small constant $\epsilon$ may be added for stabilization) the rows, so that we end up with \textbf{classification probabilities}.
\par Previously, we developed 
a semi-supervised leave-one-out filter \cite{AFONSO2020PRL,BKthesis2020}. We also managed to reduce the amount of storage used by only calculating a \textbf{propagation submatrix}. The LOO-inspired criterion encourages label information to be redundant, so labels that are incoherent with the implicit model are removed.
\par 
The major drawback of  
our proposal is that it needs an extra parameter $r$, which is the number of labels to remove. The optimal $r$ is usually around the number of noisy labels, which is unknown to us.
 This was somewhat addressed in \cite{AFONSO2020PRL}: we can instead use a threshold, which tells us how much labels can deviate from the original model. Nonetheless, it is desirable to solve this problem in a way that removes such a parameter. We will do this by introducing a new optimization problem.

\section{Proposal} \label{sec:proposal}
In this work, we consider the optimal value of ${\alpha}$ for our LOO-inspired loss \cite{AFONSO2020PRL}. The objective is to develop a method to minimize \textbf{surrogate losses} based on LOO error for the LGC GSSL classifier and evaluate the generalization of its solutions.

We use the term ``surrogate loss''\cite{krijthe2018robust} to denote losses such as squared error, cross-entropy and so on. We use this term to purposefully remind that the solution $\mb{H}(\alpha)$ that minimizes the loss on the test set does not also necessarily be the one that maximizes accuracy. We use \textbf{automatic differentiation} as our optimization procedure. 
As such, we call our approach $\LGCLVOAutoL$ and $\LGCLVOAutoD$ when learning label reliability and diffusion rate, respectively. We exploit the fact that we can compute the gradients of our loss. 

\subsection{$\LGCLVOAutoL$: Automatic Correction of Noisy Labels}
\label{sec:lgclvo_auto_correction}
\par Let $\mb{P}$ be the propagation matrix, whose submatrix $\mb{P}_{\lab\lab}$ corresponds to kernel values between labeled data only. Then, the modified LGC solution is given by $\widetilde{\mb{P}}\mb{Y}_\lab$, where
\begin{equation}
         \widetilde{\mb{P}} := (\mb{P}_{\lab\lab} -diag(\mb{P}_{\lab\lab}))
\end{equation}
Our optimization problem is to optimize a diagonal matrix $\mb{\Omega}$ indicating the updated reliability of each label:

\begin{align}
\label{eqn:mse}
    \min_{\mb{\Omega}} \texttt{LOSS}( { deg(\widetilde{\mb{P}}\mb{\Omega}\mb{Y}_\lab)^{-1} \widetilde{\mb{P}}\mb{\Omega}\mb{Y}_\lab ,\;\;\mb{Y}_\lab}) \;\;\;\;\;\;\;
    \text{such that } \;\; \forall i:  \mb{\Omega}_{ii} \geq 0
\end{align}
where $\texttt{LOSS}$ denotes some loss function such as squared error or cross-entropy.

\subsection{$\LGCLVOAutoD$: Automatic choice of diffusion rate}
In \cite{AFONSO2020PRL},  we use a modified version of the power method to calculate the submatrix $\mb{P}_\lab$. This is enough to give us the answer in a few seconds for a fixed $\alpha$, but would quickly turn into a huge bottleneck if we were to constantly update $\alpha$. Our new approach uses the graph Fourier transform. In other words, we adapt the idea of smooth eigenbasis pursuit to this particular problem. 
Let us write the propagation matrix using eigenfunctions:
\begin{align}
    \mb{P} &= \sum_{t=0}^\infty \alpha^t \mb{S}^t 
    \\& =  \sum_{t=0}^\infty \alpha^t (\mb{I}-\mb{\Ln})^t
\end{align}
Using the graph Fourier transform, we have that
\begin{align}
    \mb{\Ln} &= \mb{U}\mb{\Lambda}\mb{U}^\top
    \\\mb{I} &= \mb{U}\mb{I}\mb{U}^\top
\end{align}
So it follows that
\begin{align}
    \mb{P} &= \mb{U} \widetilde{\Lambda}\mb{U}^\top
\end{align}
where $\forall i \in \{1..N\}$:
\begin{align}
     \widetilde{\Lambda}_{ii} &= \sum_{t=0}^\infty (\alpha (1-\Lambda_{ii}) )^t 
    \\\{ \norm{\alpha (1-\Lambda_{ii}) } < 1\} &= \frac{1}{1- (\alpha (1-\Lambda_{ii}) )} 
    \\&= \frac{1}{(1 - \alpha)\times 1 + (\alpha) \times \Lambda_{ii}} 
\end{align}
In practice, we can assume that $\mb{U}$ is an $l \by p$ matrix, with $l$ the number of labeled instances and $p$ the chosen amount of eigenfunctions. The diagonal entries are given by
\begin{align}
    \mb{P}_{ii} &= (\mb{U}\widetilde{\Lambda}\mb{U}^\top)_{ii} \\&= (\mb{U}^\top)_{[:,i]}\widetilde{\Lambda}(\mb{U}^\top)_{[:,i]} 
    \\&= \Sigma_{i=1}^p \mb{U}_{ik}^2 \widetilde{\Lambda}_{kk} \\&= (\mb{U}_{[i,:]}) (\mb{U}_{[i,:]} \widetilde{\Lambda})^\top
\end{align}
Next, we'll analyze the complexity of calculating the leave-one-out error for a given diffusion rate $\alpha$, given that we have stored the first $p$ eigenfunctions in the matrix $\mb{U}$. Let $\mb{U}_\lab \in \mathbb{R}^{p \by l}$ be the matrix of eigenfunctions with domain restricted to labeled instances. Exploiting the fact that $\mb{Y}_\unlab = 0$, it can be shown that
\begin{equation}
    \mb{P}_{\lab\lab} = \mb{U}_{\lab}\mb{\widetilde{\Lambda}}(\mb{U}_{\lab}^\top \mb{Y}_\lab)
\end{equation}
We  can precompute $(\mb{U}_{\lab}^\top \mb{Y}_\lab) \in \mathbb{R}^{p \by c}$ with $\mathcal{O}(plc)$ multiplications. For each diffusion rate candidate $\alpha$, we obtain $\mb{U}_{\lab}\mb{\widetilde{\Lambda}}  \in \mathbb{R}^{l \by p}$ by multiplying each restricted eigenfunction with its new eigenvalue. The only thing left is to post-multiply it by the pre-computed matrix. As a result, we can compute the leave-one-out error for arbitrary $\alpha$ with $\mathcal{O}(plc)$ operations.
\par We have shown that, by using the propagation submatrix, we can re-compute the propagation submatrix $\mb{P}_{\lab\lab}$ in $\mathcal{O}(plc)$ time. In comparison, the previous approach \cite{AFONSO2020PRL} requires, for each diffusion rate, a total of $O(tknlc)$ operations, assuming $t$ iterations of the power method and a sparse affinity matrix with average node degree equal to $k$.  Even if we use the full eigendecomposition ($p = n$), this new approach is significantly more viable for different learning rates. Moreover, we can lower the choice of $p$ to use a faster, less accurate approximation.  
\section{Methodology}\label{sec:methodology}
\subsubsection{Basic Framework}
We have a \textbf{configuration dispatcher} which enables us to vary a set of parameters (for example, the chosen dataset and the parameters for affinity matrix generation).
\par We start out by reading our dataset, including features and labels. We use the random seed to select the sampled labels, and create the affinity matrix $\mb{W}$ necessary for LGC. For automatic label correction, we assume that $\alpha$ is given as a parameter. For choosing the automatic diffusion rate, we need to extract the $p$ smoothest eigenfunctions as a pre-processing step. Next, we repeatedly calculate the gradient and update either  $\alpha$ or $\mb{\Omega}$ to minimize LOO error. After a set amount of iterations, we return the final classification and perform an evaluation on unlabeled examples. 

The programming language of choice is \texttt{Python 3}, for its versatility and support. We also make use of the \textit{tensorflow-gpu} \cite{tensorflow2015-whitepaper} package, which massively speeds up our calculations and also enables automatic differentiation of loss functions. We use a Geforce GTX 1070 GPU to speed up inference, and also for calculating the k-nearest neighbors of each instance with the \textit{faiss-gpu} package \cite{johnson2019billion}.

\subsubsection{Evaluation and baselines}
In this work, our datasets have a roughly equal number of labels for each class. As such, we will report the mean accuracy, as well as its standard deviation. However, one distinction is that we calculate the accuracy independently on labeled and unlabeled data. This is done to better assess whether our algorithms are improving classification on instances outside the labeled set, or if it outperforms its LGC baseline only when performing diagnosis of labels. \par Our approach, $\LGCLVOAuto$, is compatible with any differentiable loss function, such as \textit{mean squared error} (MSE) or \textit{cross-entropy} (xent). The chosen optimizer was Adam, with a learning rate of 0.7 and 5000 iterations. For the approximation of the propagation matrix in $\LGCLVOAutoL$, we used $t=1000$ iterations throughout.

Perhaps the most interesting classifier to compare our approach to is the LGC algorithm, as that is the starting point and the backbone of our own approach. We will also be comparing our results with the ones reported by \cite{gong2017learning}. These include: Gaussian Fields and Harmonic Functions (GFHF) \cite{Zhu_2003}, Graph Trend Filtering \cite{wang2016trend}, Large-Scale Sparse Coding (LSSC) \cite{lu2015noise} and Eigenfunction \cite{fergus2009semi}.

\section{Results}\label{sec:results}
This section presents the results of employing our two approaches $\LGCLVOAutoL$ and $\LGCLVOAutoD$ on ISOLET and MNIST datasets, compared to other graph-based SSL algorithms from the literature.

\subsection{Experiment 1: $\LGCLVOAutoL$ on ISOLET}
\label{sec:exp1}
\subsubsection{Experiment setting}
In this experiment, we compared $\LGCLVOAutoL$ to the baselines reported in \cite{gong2017learning}, specifically for the ISOLET dataset. Unlike the authors, our 20 different seeds also control both the \textbf{label selection and noise processes}. The graph construction was performed exactly as in \cite{gong2017learning}, a symmetric 10-nearest neighbors graph with the width $\sigma$ of the RBF kernel set to 100. We emphasize that the reported results by the authors correspond to the best-performing parameters, divided for each individual noise level. 
 In \cite{gong2017learning}, $\lambda_1$ is set to  $10^5$, $10^2$, $10^2$ and $10^2$ for the respective noise rates of 0\%, 20\%, 40\% and 60\%; $\lambda_2$ is kept to 10, and the number of eigenfunctions is $m=30$. We could not find  any implementation code for SIIS, so we had to manually reproduce it ourselves. As for parameter selection for $\LGCLVOAutoL$, we simply set $\alpha = 0.9$ (equivalently, $\mu=0.1111$). We reiterate that \textbf{having a single parameter} is a \textbf{strength of our approach}. In future work, we will try to combine $\LGCLVOAutoL$ with $\LGCLVOAutoD$ to fully eliminate the need for parameter selection.
\subsubsection{Experiment results}
The results are contained in Tables  \ref{tbl:isolet_unlabeled_acc} and \ref{tbl:isolet_labeled_acc}. With respect to the accuracy on \textbf{unlabeled examples}, we observed that:
\begin{itemize}
\itemsep0em
    \item SIIS appeared to have a slight edge in the noiseless scenario.
    \item \textbf{LGC's own inherent robustness was evident}. When 60\% noise was injected, it went from $84.72$ to $70.69$, a decrease of $16.55\%$. In comparison, SIIS had a decrease of $14.39\%$; GFHF a decrease of $22.08\%$; GTF a decrease of $21.82\%$.
    \item With $60\% $ noise,  $\LGCLVOAutoL$\texttt{(xent)} decreased its accuracy by  $11.52\%$, so  \textbf{$\LGCLVOAutoL$(xent) had the lowest percentual decrease}.
    \item \textbf{$\LGCLVOAutoL$(MSE) disappointed} for both labeled and unlabeled instances. 
    \item \textbf{$\LGCLVOAutoL$ was not noticeably superior to LGC when there was less than 60\% noise}. 
\end{itemize}

With respect to the accuracy on \textbf{unlabeled examples}, we observed that:
\begin{itemize}
\itemsep0em
    \item \textbf{LGC was unable to correct noisy labels}.
    \item $\LGCLVOAutoL (xent)$ discarded only 5\% of the labels for the noiseless scenario, which is \textbf{better than SIIS and LSSC}.
    \item Moreover, \textbf{$\LGCLVOAutoL (xent)$} had the \textbf{highest average accuracy on labeled instances} for $20\%,40\%,60\%$ label noise.
\end{itemize}

\begin{table}[!h]
\begin{subtable}{\textwidth}
\centering
\scriptsize
\caption{Accuracy on \textbf{unlabeled examples} only}

\begin{tabular}{llllll}
\hline
Dataset & Noise Level & LSSC           & GTF                                        & GFHF                     & SIIS  \\\hline
                            & 0\%         & 84.8 $\pm$ 0.0   & 70.1 $\pm$ 0.0       & 86.5 $\pm$ 0.0            & \textbf{85.4 $\pm$ 0.0}     \\
\hfil ISOLET                & 20\%        & 82.8 $\pm$ 0.3   & 69.9 $\pm$ 0.2       & 81.6 $\pm$ 0.4            & \textbf{84.9 $\pm$ 0.6} \\
\hfil (1040/7797 labels)    & 40\%        & 78.5 $\pm$ 0.6   & 59.8 $\pm$ 0.3       & 79.7 $\pm$ 1.0            & \textbf{80.2 $\pm$ 1.3} \\
\hfil reported results      & 60 \%       & 67.5 $\pm$ 1.8   & 54.8 $\pm$ 0.5       & 67.4 $\pm$ 1.5            & \textbf{74.9 $\pm$ 1.4}\\ \hline
Dataset & Noise Level & LGC            & \LGCLVOAutoL (MSE)                 & \LGCLVOAutoL (XENT)                       &  SIIS\\    \hline
                            & 0\%         & 84.71 $\pm$ 0.56  & 84.21$\pm$0.4           &  84.22$\pm$0.45                   &  \textbf{85.24 $\pm$ 0.32}             \\
\hfil ISOLET                & 20\%        & 82.89 $\pm$ 0.59  & 81.6$\pm$0.63           &  82.56$\pm$0.62                   &  \textbf{83.69 $\pm$ 0.33}            \\
\hfil (1040/7797 labels)    & 40\%        & 79.33 $\pm$ 0.92  & 77.73$\pm$0.96          & 80.23$\pm$0.74                     &  \textbf{80.88 $\pm$ 0.77}          \\
\hfil our results     & 60\%              & 70.69 $\pm$ 1.01  & 68.98$\pm$1.81 &  \textbf{74.51$\pm$1.75}                &  72.97 $\pm$ 1.16
\end{tabular}
\label{tbl:isolet_unlabeled_acc}
\end{subtable}
\begin{subtable}{\textwidth}
\tiny
\caption{Accuracy on \textbf{labeled examples} after label correction}
\centering

\scriptsize
\begin{tabular}{llllll}
\hline
Dataset & Noise Level                            & LSSC           & GTF                    & GFHF               & SIIS                                  \\\hline
                            & 0\%         & 89.9 $\pm$ 0.0   & 95.8 $\pm$ 0.0       &\textbf{100.00 $\pm$ 0.0}        & 91.1 $\pm$ 0.0  \\
\hfil ISOLET                    & 20\%        & 87.7 $\pm$ 0.3   & 79.8 $\pm$ 0.7           &80.00 $\pm$ 0.00        & \textbf{90.5 $\pm$ 0.8}  \\
\hfil (1040/7797 labels)        & 40\%        & 82.9 $\pm$ 0.9   & 63.3 $\pm$ 0.4           &60.00 $\pm$ 0.00        & \textbf{83.6 $\pm$ 1.0}  \\
\hfil reported results                & 60 \%       & 71.8 $\pm$ 1.7   & 55.3 $\pm$ 0.6     &40.00 $\pm$ 0.00        & \textbf{77.4 $\pm$ 1.0} \\ \hline
Dataset & Noise Level               & LGC                           & \LGCLVOAutoL (MSE)                  &\LGCLVOAutoL (XENT)                 & SIIS       \\    \hline
                    & 0\%         & \textbf{99.9 $\pm$ 0.02}          &  97.36$\pm$0.52           &   95.01$\pm$0.58   & 90.24 $\pm$ 0.69      \\
\hfil ISOLET                    & 20\%        & 80.84 $\pm$ 0.27      &  90.93$\pm$1.28         &    \textbf{91.52$\pm$0.79}  & 88.5 $\pm$ 1.07  \\
\hfil (1040/7797 labels)        & 40\%        & 60.24 $\pm$ 0.20      &  82.54 $\pm$ 0.92       &    \textbf{87.19$\pm$0.89}    & 85.25 $\pm$ 0.96    \\
\hfil our results                     & 60\%  & 40.00 $\pm$ 0.04& 71.16 $\pm$ 1.79     &     \textbf{79.14$\pm$1.72}     & 76.34 $\pm$ 1.53   
\end{tabular}
\label{tbl:isolet_labeled_acc}
\end{subtable}
\caption{Accuracy on \texttt{ISOLET} dataset}
\end{table}

\subsection{Experiment 2: $\LGCLVOAutoL$ on MNIST}
\label{sec:exp2}
\subsubsection{Experiment setting}
This experiment was based on \cite{lu2015noise}, where a few classifiers were tested on the MNIST dataset subject to label noise. In that paper, the parameters for the graph were tuned to minimize cross-validation errors. Moreover, an anchor graph was used, which is a large-scale solution. We did not use such a graph, as \textbf{our TensorFlow iterative implementation of $\LGCLVOAutoL$ was efficient enough to perform classification on MNIST in just a few seconds}.  As we also included the results for LGC (without anchor graph), it is interesting to observe that its accuracy decreases similarly to the previously reported results: the main difference is better performance for the noiseless scenario, which is to be expected (the anchor graph is an approximation).
\par Once again, we simply set $\alpha=0.9$ for $\LGCLVOAutoL$. We used a symKNN matrix with $k=15$ neighbors, and a heuristic sigma $\sigma=423.57$ obtained by taking one-third of the mean distance to the 10th neighbor (as in \cite{Chapelle_etal_2010}).

\begin{table}[!h]
\begin{subtable}{\textwidth}
\centering
\scriptsize
\caption{Accuracy on \textbf{unlabeled examples} only}

\begin{tabular}{llllll}
\hline
Dataset                     & Noise Level & LSSC*             & Eigenfunction*            & LGC* (anchor graph)                    &                                   \\\hline
\hfil MNIST                 & 0\%         & \textbf{93.1 $\pm$ 0.7}   &73.8 $\pm$ 1.6            & 90.4 $\pm$ 0.7                       &  \\
\hfil (100/70000 labels)    & 15\%        & \textbf{91.1 $\pm$ 2.0}   &68.6 $\pm$ 2.8            & 83.5 $\pm$ 1.6                      & \\
\hfil reported results      & 30\%        & \textbf{89.0 $\pm$ 3.6}   &61.9 $\pm$ 4.0            & 74.4 $\pm$ 2.8                      & \\ \hline

Dataset                      & Noise Level & \LGCLVOAutoL (MSE)            &\LGCLVOAutoL (XENT)                  &LGC                        &  \\    \hline
\hfil MNIST                 & 0\%         & 91.7 $\pm$ 0.7      & 92.69 $\pm$ 1.19           &\textbf{93.09 $\pm$ 0.92}            &  \\
\hfil (100/70000 labels)    & 15\%        & 86.48 $\pm$ 2.59   & \textbf{90.45 $\pm$ 2.13}           &85.40 $\pm$ 1.66           & \\
\hfil our results           & 30\%        & 81.33 $\pm$ 4.43   & \textbf{84.46 $\pm$ 3.89}           &74.58 $\pm$ 2.6          & \\
\end{tabular}
\label{tbl:mnist_unlabeled_acc}
\end{subtable}
\begin{subtable}{\textwidth}
\tiny
\caption{Accuracy on \textbf{labeled examples} after label correction}
\centering

\scriptsize
\begin{tabular}{llllll}

Dataset                      & Noise Level & \LGCLVOAutoL (MSE)            &\LGCLVOAutoL (XENT)                  &LGC                        &  \\    \hline
\hfil MNIST                 & 0\%         & 99.5 $\pm$ 0.59     & 98.05 $\pm$ 0.59                  &\textbf{100.00 $\pm$ 0.0}            &  \\
\hfil (100/70000 labels)    & 15\%        & 95.05 $\pm$ 2.01   & \textbf{96.10 $\pm$ 1.3}           &85.00 $\pm$ 0.0           & \\
\hfil our results           & 30\%        & 85.55$\pm$4.88   & \textbf{89.75 $\pm$ 4.06}           &70.00 $\pm$ 0.0          & \\
\end{tabular}
\label{tbl:mnist_labeled_acc}
\end{subtable}
\caption{Accuracy on \texttt{MNIST} dataset}
\end{table}

\subsubsection{Experiment results}

The results are found in Tables \ref{tbl:mnist_unlabeled_acc} and \ref{tbl:mnist_labeled_acc}. With respect to the accuracy on \textbf{unlabeled examples}, we observed that:
\begin{itemize}
\itemsep0em
    \item \textbf{\LGCLVOAutoL with cross-entropy improved the LGC baseline significantly on unlabeled instances}. For 30\% label noise, mean accuracy increases \textbf{from 74.58\% to 84.46\%}.
    \item The \textbf{mean squared error loss is once again consistently inferior to cross-entropy when there is noise}.
    \item $\LGCLVOAutoL$ was not able to obtain better results than LSSC. 
\end{itemize}

With respect to the accuracy on \textbf{labeled examples}, we observed that:
\begin{itemize}
\itemsep0em
    \item \textbf{LGC was not able to correct the labeled instances}.
    \item \textbf{\LGCLVOAutoL with cross-entropy improved the LGC baseline significantly on labeled instances}. With 30\% noise, accuracy goes \textbf{from 70.00\% to 89.75\%}. 
\end{itemize}

\subsection{Experiment 3: $\LGCLVOAutoD$ on MNIST}
\begin{figure}[!ht]
     \centering
     \begin{subfigure}[b]{0.45\textwidth}
         \centering
         \includegraphics[width=\textwidth]{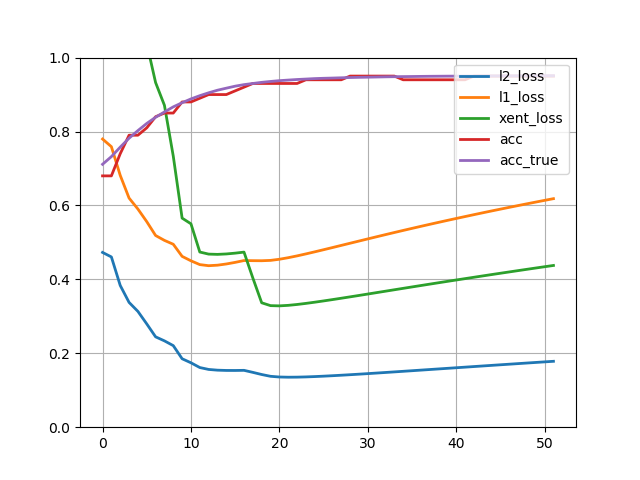}
         \caption{Diagonal removed}
         \label{fig:lgclvod_RE}
     \end{subfigure}
     \hfill
     \hfill
     \begin{subfigure}[b]{0.45\textwidth}
         \centering
         \includegraphics[width=\textwidth]{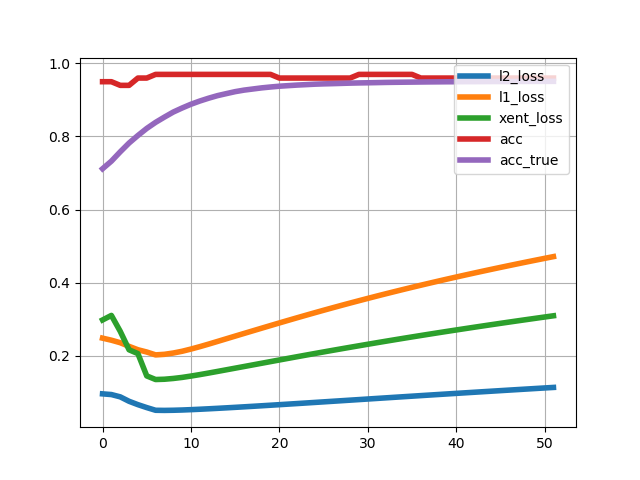}
         \caption{Diagonal NOT removed}
         \label{fig:lgclvod_NR}
     \end{subfigure}
        \caption{$\LGCLVOAutoD$ on MNIST.  The $x$ coordinate on the horizontal axis determines diffusion rate $\alpha = 2^{-1/x}$ for $x\geq 1$. The orange, blue and green curves represent the surrogate losses: {mean absolute error}, {mean squared error},   {cross-entropy}. Moreover, the accuracy on the known labeled data is shown in red, and the accuracy on unlabeled data in purple.}
        \label{fig:lgclvod}
\end{figure}

Experiment settings were kept the same as Experiment 2 (without noise), and $p = 300$ eigenfunctions were extracted. In Figure \ref{fig:lgclvod}, we show accuracy on the labeled set as red, and on the unlabeled set as purple. The $x$ values on the horizontal axis relate to $\alpha$ as following: $\alpha = 2^{-1/x}$. Looking at Figure \ref{fig:lgclvod_NR}, we can see that, if we do not remove the diagonal, there is much overfitting and the losses reach their minimum value much earlier than does the accuracy on unlabeled examples. When the diagonal is removed, the loss-minimizing estimates for $\alpha$ get much closer to the optimal one (Figure \ref{fig:lgclvod_RE}). Moreover, by removing the diagonal we obtain a much better estimation of the accuracy on unlabeled data. Therefore, we can simply select the $\alpha$ corresponding to the best accuracy on known labels.

\section{Concluding remarks}\label{sec:conclusion}
We have proposed the $\LGCLVOAuto$ framework, based on leave-one-out validation of the LGC algorithm. This encompasses two methods, $\LGCLVOAutoL$ and $\LGCLVOAutoD$, for estimating label reliability and diffusion rates. We use automatic differentiation for parameter estimation, and the eigenfunction approximation is used to derive a faster solution for $\LGCLVOAutoD$, in particular when having to recalculate the propagation matrix for different diffusion rates.
\par Overall, $\LGCLVOAutoL$ produced interesting results. For the ISOLET dataset, it was \textbf{very successful at the task of label diagnosis}, being able to detect and remove labels with overall better performance than every $\ell_1$-norm method. On the other hand, \textbf{did not translate too well for unlabeled instance classification}. For the MNIST dataset, \textbf{performance is massively boosted for unlabeled instances as well}. In spite of outperforming its LGC baseline by a wide margin, it could not match the reported results of LSSC on MNIST. Preliminary results showed that \textbf{$\LGCLVOAutoD$} is a viable way to get a good estimate of the optimal diffusion rate, and removing the diagonal entries proved to be the crucial step for avoiding overfitting.
\par For future work, we will be further evaluating $\LGCLVOAutoD$. We will also try to integrate $\LGCLVOAutoD$ and $\LGCLVOAutoL$ together into one single algorithm. Lastly, we will aim to extend $\LGCLVOAutoD$ to a broader class of graph-based kernels, in addition to the one resulting from the LGC baseline.
\bibliographystyle{splncs04}
\bibliography{mybibliography}

\end{document}